\documentclass[10pt,twocolumn,letterpaper]{article}

\usepackage[pagenumbers]{cvpr} % To force page numbers, e.g. for an arXiv version

\definecolor{cvprblue}{rgb}{0.21,0.49,0.74}
\usepackage[pagebackref,breaklinks,colorlinks,allcolors=cvprblue]{hyperref}
\usepackage{bm}
\usepackage{amsmath}
\usepackage{multirow}
\usepackage{indentfirst}
\usepackage{xcolor}
\usepackage{colortbl}
\def\paperID{*****} % *** Enter the Paper ID here
\def\confName{CVPR}
\def\confYear{2025}

\title{Similarity-Guided Layer-Adaptive Vision Transformer for UAV Tracking}

\author{Chaocan Xue\textsuperscript{1}, Bineng Zhong\textsuperscript{1}\thanks{Corresponding Author}, Qihua Liang\textsuperscript{1}\thanks{Corresponding Author}, Yaozong Zheng\textsuperscript{1}, Ning Li\textsuperscript{1} \\
Yuanliang Xue\textsuperscript{2}, Shuxiang Song\textsuperscript{1} \\
\textsuperscript{1}Key Laboratory of Education Blockchain and Intelligent Technology,\\
Ministry of Education, Guangxi Normal University, Guilin 541004, China\\
\textsuperscript{2}Xi'an Research Institute of High Technology, Xi'an 710025, China\\
{\tt\small xcc23cg@163.com, bnzhong@gxnu.edu.cn, qhliang@gxnu.edu.cn}\\ 
{\tt\small yaozongzheng@stu.gxnu.edu.cn, ningli65536@mailbox.gxnu.edu.cn } \\
{\tt\small xyl\_507@outlook.com, songshuxiang@mailbox.gxnu.edu.cn }
}

\begin{document}
\maketitle
\begin{abstract}
Vision transformers (ViTs) have emerged as a popular backbone for visual tracking. However, complete ViT architectures are too cumbersome to deploy for unmanned aerial vehicle (UAV) tracking which extremely emphasizes efficiency. In this study, we discover that many layers within lightweight ViT-based trackers tend to learn relatively redundant and repetitive target representations. Based on this observation, we propose a similarity-guided layer adaptation approach to optimize the structure of ViTs. Our approach dynamically disables a large number of representation-similar layers and selectively retains only a single optimal layer among them, aiming to achieve a better accuracy-speed trade-off. By incorporating this approach into existing ViTs, we tailor previously complete ViT architectures into an efficient similarity-guided layer-adaptive framework, namely SGLATrack, for real-time UAV tracking. Extensive experiments on six tracking benchmarks verify the effectiveness of the proposed approach, and show that our SGLATrack achieves a state-of-the-art real-time speed while maintaining competitive tracking precision. Codes and models are available at https://github.com/GXNU-ZhongLab/SGLATrack.

\end{abstract}     
\section{Introduction}
\label{sec:intro}

\begin{figure}[t]
  \centering
   \includegraphics[width=3.2in]{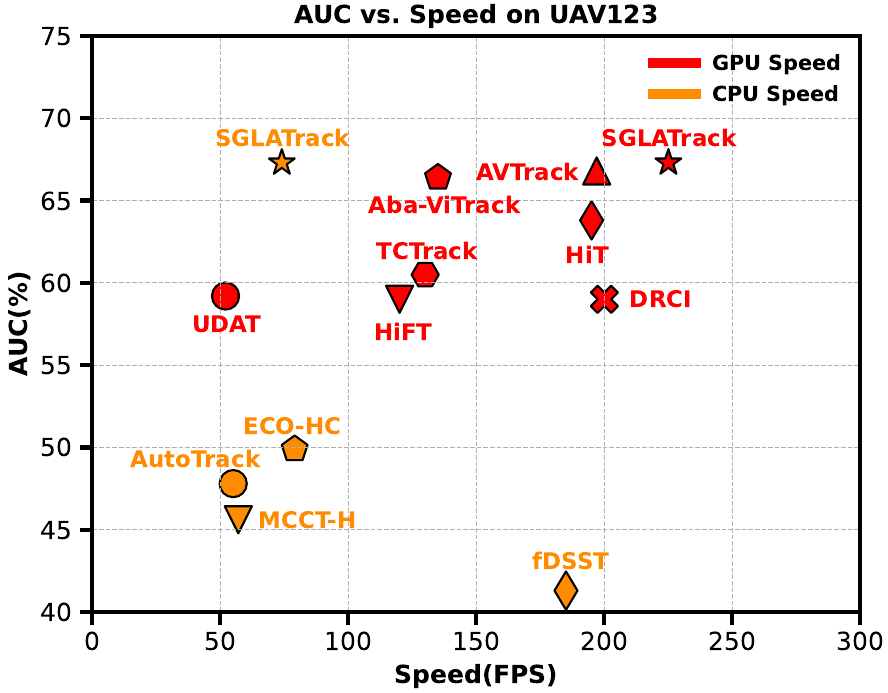}
   \caption{Comparison of SGLATrack against other UAV trackers on UAV123. Our SGLATrack demonstrates state-of-the-art performance with an AUC score of 66.9\%, while running efficiently at nearly 225 FPS on GPU and approximately 75 FPS on CPU.}
   \label{fig:fps}
\end{figure}

Unmanned Aerial Vehicle (UAV) tracking is an essential branch of visual object tracking (VOT), which can be used for various applications such as visual surveillance \cite{Autotrack}, path planning \cite{Intro1} and border security \cite{Intro2}. However, compared to generic object tracking, UAV tracking typically faces greater challenges due to its unique aerial perspective. More importantly, the limited power and computing resources of UAVs impose strict demands on efficiency. Therefore, an excellent UAV tracker must maintain both high tracking precision and efficiency.

Owing to the high efficiency derived from the operations in the Fourier domain, discriminative correlation filter (DCF)-based methods are widely used for UAV Tracking. Despite the efficiency, they hardly achieve favorable tracking robustness using hand-crafted feature extractors \cite{KCF,DSST,SRDCF}. Siamese network (SN)-based trackers can learn robust target representations using deep features, but their reliance on early convolutional neural networks (CNNs) leaves considerable potential for precision improvement \cite{SiamFC,DaSiamRPN,SiamRPN}. In recent years, vision transformer (ViT)-based trackers have become increasingly popular \cite{transt,Stark,Mixformer,OSTrack}. Notable examples are Mixformer 
 \cite{Mixformer} and OSTrack \cite{OSTrack}, which drive the shift of tracking models towards highly parallel one-stream architectures. Nonetheless, they are still too cumbersome to be directly deployed for real-time UAV tracking. To improve the efficiency, Aba-ViTrack \cite{ABA} introduces an adaptive and background-aware token halting approach into tiny ViTs. Also based on tiny ViTs, AVTrack \cite{AVTrack} proposes a more structured approach that dynamically enables or disables ViT layers according to input complexity, further improving efficiency by avoiding extra unstructured access operations. However, such an approach requires attaching classifiers to most ViT layers to decide whether each layer should be activated, which is time-consuming and redundant. Moreover, although their works demonstrate the potential of tiny ViTs in UAV tracking tasks, a surprising finding is that the layer redundancy in lightweight ViT-based trackers remains unexplored. 
 
 % although their acceleration for lightweight ViTs has significantly promoted the development of UAV tracking, a surprising finding is that the layer redundancy within lightweight ViT-based trackers remains unexplored.

% although the above trackers have greatly advanced ViT-based drone tracking, the redundancy of ViT has remained unexplored.

% such a approach relies on the output confidences of internal classifiers as a proxy for input complexity, but these outputs are not always reliable, especially for lightweight classifiers. Moreover, attaching classifiers to each ViT layer is redundant, and ViT layers may be distracted by the conflicting tasks of training excessive classifiers.

In this paper, we make the first attempt to study the layer redundancy within lightweight ViT-based trackers, with the goal of providing more advanced guidelines for accelerating ViT-based UAV tracking. Our intuition is based on the saturation observation \cite{observation1,observation2}, which indicates the hidden states of transformers reach saturation as they progress into deeper layers. For visual tracking tasks, we demonstrate that a similar saturation conclusion is also applicable. Specifically, we provide a set of empirical experiments to study the layer redundancy of tiny ViTs, including layer-by-layer cosine similarity analysis and a check of the performance impact when stopping inference at a specific layer. These studies show that the encoded search features undergo significant changes in shallow layers, but will reach a saturation state at a certain layer. Once such a saturated layer is reached, the features exhibit relatively small changes when passing subsequent layers, and the impact of these subsequent layers on the final predicted results also decreases. Therefore, some subsequent layers can be disabled without causing significant performance degradation. To this end, we propose a similarity-guided layer adaptation approach, which disables a large number of these layers and selectively retains only a single representative layer among them, aiming to achieve a better accuracy-speed trade-off. Our approach is implemented with a selection module, which is a simple multi-layer perceptron (MLP). It takes partial features at the saturated layer as input, and outputs the probabilities of each subsequent layer being selected. Considering that each layer learns different representations and that different tracking scenarios require different representations, we introduce a layer-wise similarity loss to optimize the selection of this module, allowing the model to dynamically maximize its focus on the target in situations where a large number of layers are disabled. By incorporating this approach
into existing ViTs, we tailor a family of efficient trackers, namely SGLATrack, to meet the real-time demands of UAV Tracking. Extensive experiments on six popular datasets demonstrate that SGLATrack achieves a better accuracy-speed trade-off. Notably, as illustrated in \cref{fig:fps}, SGLATrack achieves  remarkable performance of 66.9\% AUC while maintaining an impressive tracking speed of 225 frames per second (FPS) on UAV123. Our main contribution can be summarized as follows:
\begin{itemize}
\item We conduct a thorough analysis of layer redundancy in lightweight ViT-based trackers by examining layer-by-layer feature changes and result variations. Our analysis is meticulous and provides valuable insights.

\item We propose a similarity-guided layer adaptation approach to optimize redundant ViT layers. This approach can be seamlessly integrated into existing ViTs, and achieve a better accuracy-speed trade-off.

\item We develop SGLATrack, a family of efficient trackers based on different optimized ViTs for real-time UAV tracking. SGLATrack achieves state-of-the-art performance on six challenging benchmarks, while setting a new record of tracking speed.
\end{itemize}

% Considering that the saturated features already contain some knowledge of the target, we propose a layer-wise similarity loss to guide these features into specific subsequent layers, allowing them to quickly focus on the target and enabling the model to complete learning earlier.

% casts object detection as a task of generating tokens based on observed pixel inputs, driving the shift of tracking tasks towards the paradigm of sequence generation.

%  In recent years, deep learning (DL)-based trackers have become increasingly popular.

%-------------------------------------------------------------------------

\section{Related Works}
\label{sec:formatting}

\subsection{Visual Object Tracking}
Mainstream VOT algorithms can be roughly divided into three types: DCF-based , SN-based, and ViT-based methods. DCF-based trackers are widely used in the field of UAV tracking for their high efficiency, but they hardly maintain robustness under challenging conditions due to poor representation ability of hand-crafted features \cite{Mosse,KCF,DSST,CCOT,ECO,Autotrack,ARCF}. SN-based trackers can learn robust target representations using deep features, but most of them still rely on early CNNs, leaving precision improvement far away from being fully explored \cite{SiamFC,DaSiamRPN,SiamRPN,SiamDW,Siamfc++,SiamBan}. Recently, ViTs have been successfully introduced into visual tracking \cite{FFT,TemTrack,RGBT1,EVP,AQA,STT}. For example, Mixformer \cite{Mixformer} and OSTrack \cite{OSTrack} integrate feature extraction and interaction into a same encoder, allowing a more full fusion of search features with template features and establishing a highly parallelized one-stream architecture. Although these methods have significantly improved tracking accuracy, they are still too cumbersome for UAV tracking which highly emphasizes efficiency. To boost the efficiency, TATrack \cite{TATrack} propose a novel mutual information maximization-based knowledge distillation approach. DDCTrack \cite{DDCTrack} proposes a dynamic token sampler to  adjust the number of tokens in each stage of the transformer. Aba-ViTrack \cite{ABA} proposes an adaptive and background-aware token halting approach that learns to prune uninformative background tokens. Nonetheless, the variable number of tokens introduces significant time costs because of additional unstructured  assess operations. To overcome this limitation, AVTrack \cite{AVTrack} proposes activation modules that dynamically activate ViT layers according to input complexity. It evaluates the complexity of input samples and dynamically skips ViT layers based on the output probability from internal modules. However, the complexity of the input is difficult to estimate, and attaching an internal classifier to each layer is redundant. If an input sample is already deemed an easy case in shallow layers, the classifiers in deeper layers do not need to make another judgment. In this work, we conduct a thorough study of the redundancy in ViT layers and implement a more precise layer deactivation for ViTs.

\subsection{Efficient Vision Transformers}
ViTs have significantly impacted the field of computer vision with their outstanding performance \cite{DETR,VIT,DEIT}. Despite their widely recognized modeling capabilities, they have faced challenges related to speed limitations, especially on resource-constrained edge devices. Therefore, many works have been proposed to accelerate ViTs, primarily through low-rank approximation \cite{LRQ}, fixed pruning \cite{MD1,MD2}, and hybrid designs \cite{HY1,HY2}. However, ViTs with low-rank approximation and quantization methods often achieve efficiency gains at the cost of significant accuracy loss. Fixed pruning-based ViTs typically involve a time-consuming fine-tuning process, while hybrid ViTs struggle to flexibly adapt to different input sizes owing to their CNN-based stems. Recently, methods based on conditional computation have gained significant attention due to their ability to adaptively adjust computational resources based on input complexity. They can be mainly divided into two types: those focusing on adaptive token pruning \cite{DynamicVIT,AVIT}, and those focusing on adaptive layer activation \cite{LGVIT,VIT-EE,AVTrack}. The first type, such as DynamicViT \cite{DynamicVIT} and AViT \cite{AVIT}, designs certain rules or modules to adaptively prune tokens at different depths. Due to the additional time cost introduced by their unstructured access operations, the second type adaptively activate ViT layers according to input complexity. For instance, ViT-EE \cite{VIT-EE} proposes to add internal classifiers behind each ViT layer, and use the output of these classifiers which measures how confident the model is in the current predictions to determine whether the inference can be terminated early. Similarly, AVTrack \cite{AVTrack} attaches lightweight modules before each ViT layer to decide whether this layer needs to be executed. Despite improved inference speeds, recent works \cite{LGVIT,observation2,DEECAP} show that the input complexity is difficult to estimate and the output of the classifiers is not always reliable. The improvement in output reliability generally comes with a larger classifier, making it challenging to achieve a better accuracy-speed trade-off. In this work, we evaluate the selection of a subsequent layer only when the features are saturated, thereby avoiding excessive classifiers and assessments of input complexity.

\section{Method}
In this section, we first revisit the typical ViT-based one-stream tracking framework and present a detailed analysis of layer redundancy within this framework. Then, we illustrate how our  similarity-guided layer adaptation optimizes redundant layers. Finally, we introduce the proposed family of efficient trackers, i.e., SGLATrack.

\subsection{Layer Redundancy}
To facilitate understanding of the redundancy within ViT layers based on the saturation observation, we first offer a review of ViT-based tracking frameworks. Given a template image 
$\bm{Z} \in \mathbb{R}^{3\times H_z \times W_z}$ 
and a search image 
$\bm{S} \in \mathbb{R}^{3\times H_s \times W_s}$, 
they are first split and tokenized into a template token sequence $\bm{X}_z \in \mathbb{R}^{N_z \times D} $ and a search token sequence $\bm{X}_s \in \mathbb{R}^{N_s \times D} $ via a patch embedding layer $\mathcal{E}$. This process then results in a token sequence $\bm{X}$ consisting of $\bm{X}_z $ and $\bm{X}_s $, formulated as:

\begin{equation}
\bm{X}=[\bm{X}_z,\bm{X}_s] = \mathcal{E}(\bm{Z}, \bm{S}) \in \mathbb{R}^{N \times D},
\label{eq:PElayer}
\end{equation}
where $D$ denotes the embedding dimension, $[\cdot]$ represents the concatenation operation, $N$ is the total number of tokens and $N=N_z+N_s$. Let $\mathcal{T}^i$ denote the $i$-th ViT layer, the sequence are transformed via $\bm{X}^i=\mathcal{T}^{i}(\bm{X}^{i-1})$. Then, the operation process of standard ViTs with $l$ layers can be written as:

\begin{equation}
\bm{X}^l = \mathcal{T}^l \circ \mathcal{T}^{l-1} \dots \circ \mathcal{T}^1 \circ \mathcal{E}(\bm{Z}, \bm{S}),
\label{eq:VIT}
\end{equation}
where $\circ$ denotes the composition operation, and $\bm{X}^l$ is the final feature after the last layer which is equivalent to $[\bm{X}^{l}_z,\bm{X}^{l}_s]$. Finally, the target bounding box $\bm{B}$ can be located on the search image via $\bm{B}=\mathcal{H}(\bm{X}^{l}_s)$, where $\mathcal{H}$ is a prediction head.

\begin{figure}[t]
  \centering
   \includegraphics[width=3.25in]{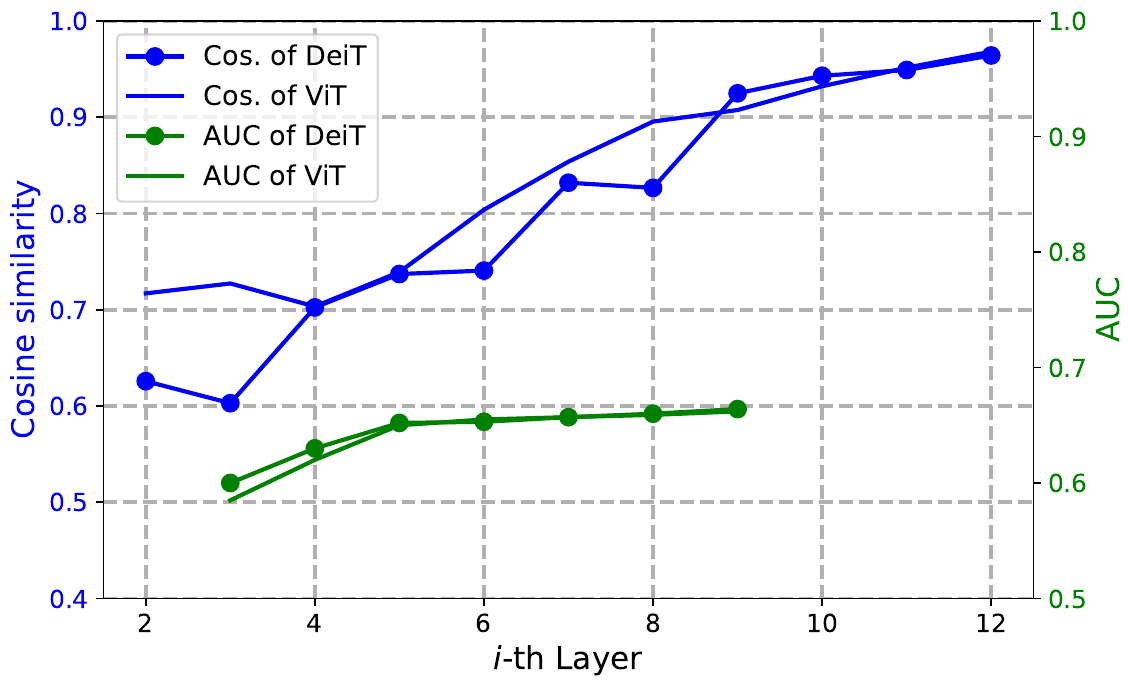}
   \caption{Layer-by-layer feature changes and AUC changes. Higher cosine similarity denotes fewer feature changes. }
   \label{fig:cos}
\end{figure}

As can be seen, the final result depends on the search features $\bm{X}^{l}_s$ and the prediction head $\mathcal{H}$. Based on this observation, a natural idea is that if the search features change barely after passing through the ViT layers, their impact on the final prediction will also be slight. This idea inspire us to analyze the changes of search features at different depths. Hence, we study whether saturation of search features exists in UAV Tracking models. Specifically, we first train several lightweight ViTs for UAV Tracking tasks, including the tiny versions of distilled DeiT \cite{DEIT} and ViT \cite{VIT} (The training settings are the same as those of OSTrack). Then, we employ cosine similarity to measure the differences between the features of each layer. Let $\textbf{a}$ and $\textbf{b}$ denote two random vectors, the cosine similarity between $\textbf{a}$ and $\textbf{b}$ is expressed as:

\begin{equation}
\text{Cos}(\textbf{a},\textbf{b}) = 
\frac{\textbf{a} \cdot \textbf{b}}{\|\textbf{a}\| \cdot \|\textbf{b}\|} ,
\label{eq:cos}
\end{equation}
where $\|\cdot\|$ represents the $l_2$ norm. \cref{fig:cos} shows the cosine similarity $\text{Cos}(\bm{X}^i_s,\bm{X}^{i-1}_s)$ between two consecutive layers and the AUC scores obtained by $\bm{B}=\mathcal{H}(\bm{X}^{i}_s)$ on UAV123 \cite{uav123}. It can be observed that the cosine similarity is higher in deep layers, indicating that the changes of search features are relatively smaller within the deeper layers. Moreover, the AUC score rapidly increases at first and then grows slowly, confirming the existence of saturation and indicating that the redundancy is more than expected. Therefore, some deep layers are redundant and can be disabled without significant precision drop, which also aligns with the consensus that shallow details are more beneficial than deep semantics for distinguishing objects in tracking tasks.

\begin{figure*}[t]
\begin{center}
\includegraphics[width=1\linewidth]{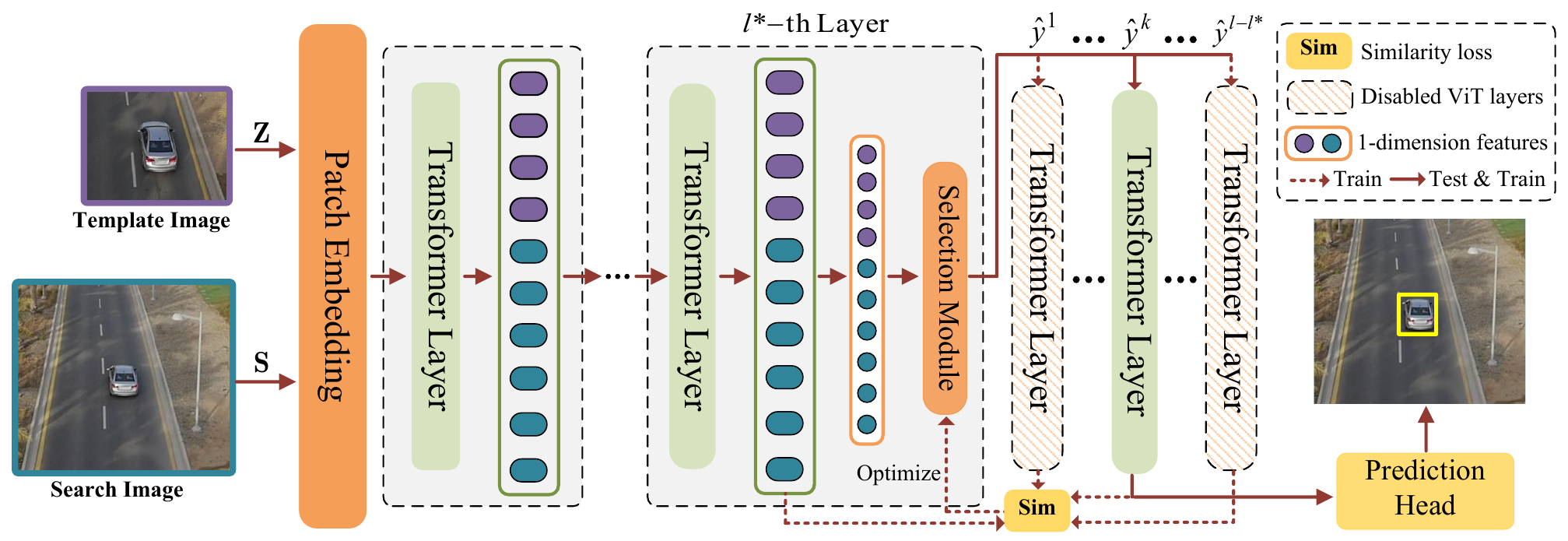}
\end{center}
   \caption{Overall architecture of the proposed SGLATrack. It is composed of a one-stream backbone, a typical prediction head, and a selection module. During training, the selection module is optimized by the proposed layer-wise similarity loss. During inference, the selection module disables redundant ViT layers and selectively retains an optimal layer among them to alleviate performance drop.}
   \label{fig:overall}
\end{figure*}

\subsection{Similarity-Guided Layer Adaptation}

According to the above observation on layer redundancy, we propose a similarity-guided layer adaptation strategy to accelerate ViT-based UAV tracking. Although deeper layers exhibit higher similarity, directly pruning them leads to a significant drop in precision, as they learn relatively repetitive representations rather than useless ones. To this end, we can retain a single representative layer, whose output should be similar to the output of the saturated layer for minimizing precision drop. The motivation behind this is that the model will gradually focus on the target as the encoder layers go deeper. If the saturated layer has already focused attention on the target, the subsequent layers will further enhance this attention by suppressing interference. Therefore, a higher similarity between the output of the subsequent layers and the output of the saturated layer is more conducive to focused and consistent attention. Let $\bm{X}^{l^*}$ denotes the saturated features at the $l^*$-th layer, our objective is to find the $(l^*+k)$-th subsequent layer that maximizes the similarity between $\bm{X}^{l^*}$ and $\bm{X}^{l^*+k}$, expressed as:
\begin{equation}
\mathop{\arg\max}\limits_{k} \; \text{Cos}(\bm{X}^{l^*},\bm{X}^{l^*+k}) ,
\label{eq:objective}
\end{equation}
where $k \leq l-l^*$ is the index of the subsequent layers after the 
$l^*$-th layer. Given that a fixed retained layer struggles to fit all cases, we develop a selection module to dynamically choose the optimal layer for different samples. This module is a simple MLP denoted as $\mathcal{M}$, it receives a portion of saturated features as input with efficiency in consideration, and outputs the probabilities of each subsequent layer being selected. Following AVTrack~\cite{AVTrack}, the input is the features on the first dimension of $\bm{X}^{l^*}$, expressed as $ \textbf{e}^T_1 \bm{X}^{l^*} := \textbf{z} \in \mathbb{R}^{N}$, where $\textbf{e}^T_1 =[1,0,\ldots,0] \in \mathbb{R}^{N}$ is a standard unit vector. And the selection module is defined as:

\begin{equation}
\hat{\bm{y}}=\sigma(\mathcal{M}(\textbf{z})) \in \mathbb{R}^K
\label{eq:module}
\end{equation}
where $\sigma(\cdot)$ is the sigmoid function, ${\hat{\bm{y}}}=[\hat{y}^1,\hat{y}^2,\ldots,\hat{y}^{K}]$ represents the selected probabilities of the layers after the $l^*$-th layer, and $K=l-l^*$. A layer with the highest probabilities will be retained while the remaining layers will be disabled. According to the objective in Eq.~\ref{eq:objective}, we introduce a layer-wise similarity loss $\mathcal{L}_{sim}$ to optimize the selection module in making the optimal choices. This loss penalizes layers with low similarity and encourages layers with high similarity, formulated by:
\begin{equation}
\mathcal{L}_{sim} = \frac{1}{K}\sum_{k=1}^{K}|\hat{y}^k-y^k|, 
\label{eq:loss}
\end{equation}
where $y^k$ is the expected probability defined by:
\begin{equation}
y^k = 
\begin{cases}
1, & \text{if } \text{Cos}(\bm{X}^{l^*},\bm{X}^{l^*+k}) \text{ is the maximum,} \\
0, & \text{otherwise.}
\end{cases}
\label{eq:loss_condition}
\end{equation}
With this loss, the selection module can dynamically allocate the optimal layers for different tracking scenarios, allowing the model to concentrate maximally on the target in situations where a large number of layers are disabled.

\subsection{SGLATrack for UAV Tracking}

\textbf{Overall Architecture.} By incorporating the proposed similarity-guided layer adaption approach into existing ViTs, we build SGLATrack, a family of efficient trackers for real-time UAV Tracking, whose overall architecture is shown in \cref{fig:overall}. On the one hand, it follows the highly parallelized one-stream tracking framework. On the other hand, it dynamically optimizes the redundant ViT layers within this framework. Specifically, SGLATrack first restructures the standard ViT in Eq.~\ref{eq:VIT} into an efficient layer-adaptive architecture, formulated as $\bm{X}^{l^*+k} = \mathcal{T}^{l^*+k} \circ \mathcal{T}^{l^*} \dots \circ \mathcal{T}^1 \circ \mathcal{E}(\bm{Z}, \bm{S})$, for efficient interaction between template images and search images. Then, SGLATrack employs a typical prediction head to estimate target location, i.e., $\bm{B}=\mathcal{H}(\bm{X}^{l^*+k}_s)$. Note that $l^*$ can be considered as a hyperparameter to balance the trade-off between precision and speed in our framework, and $k$ is decided by the selection module. We set $l^*$ to 6, and we provide detailed discussions about the determination of saturated states in the ablation study section.

\textbf{Head and Loss.} Referring to the previous work, the prediction head $\mathcal{H}$ is a center-based head for predicting the centroid position and scale of the target. The search features $\bm{X}^{l^*+k}_s$ are first turned into a 2D spatial feature map, which is then fed into prediction head $\mathcal{H}$. Suppose the size of template and search patches is $P \times P$, the outputs of $\mathcal{H}$ are a classification score map 
$\textbf{p}\in [0,1]^{H_s/P \times W_s/P}$, a normalized bounding
box size $\textbf{s}\in [0,1]^{2\times H_s/P \times W_s/P }$, and a offset $\textbf{o}\in [0,1]^{2\times H_s/P \times W_s/P }$. The location with the highest classification score is taken as the target’s location, and the final tracking result is calculated by combining the offset size and the bounding box size. We use the  focal loss \cite{FOCAL} for classification, a combination of $L_1 $loss and GIoU loss \cite{GIOU} for bounding box regression. Finally, the overall loss can be described as:
\begin{equation}
\mathcal{L}= \mathcal{L}_{cls} + \lambda_{iou}\mathcal{L}_{iou} +\lambda_{L1}\mathcal{L}_{L1} + \gamma \mathcal{L}_{sim},
\end{equation}
where $\lambda_{iou}=2$, $\lambda_{L1}=5$, and $\gamma=0.2$ are the regularization parameters.

\section{Experiments}

\subsection{Implementation Details}
\textbf{Model.} We employ different ViTs as backbones, including ViT-tiny \cite{VIT}, distilled DeiT-tiny \cite{DEIT}, and EVA-tiny \cite{EVA}, to build three trackers for extensive evaluation, i.e., SGLATrack-ViT, SGLATrack-DeiT*, and SGLATrack-EVA, respectively. The selection module $\mathcal{M}$ is a 3-layer MLP whose hidden dimension is 160, and the $l^*$ is set to 6. In terms of input data, we take a template image with 128×128 pixels and a search image with 256×256 pixels as the input to the trackers.

\begin{table*}[t]
  \centering
    \caption{AUC, precision and speed comparisons with some state-of-the-art trackers on the test set of DTB70, UAVDT, UAV123, UAVTrack112, UAVTrack112$\_$L. For the comparison between SGLATrack-Deit$^*$ and other trackers, the top two results are highlighted with {\color{red}red}, {\color{blue}blue}, respectively. For the comparison of SGLATrack using different backbones, the best results are shown in \underline{underline} font. Note that the average performance across the five datasets is presented in the form of Avg.} 
  \resizebox{\textwidth}{!}{
     \fontsize{19}{20}\selectfont
     \renewcommand{\arraystretch}{1.15}  
  \begin{tabular}{c|c|cc|cc|cc|cc|cc|cc|cc}
    \toprule
  \multicolumn{1}{c|}{\multirow{2}{*}{Method}} & \multicolumn{1}{c|}{\multirow{2}{*}{Source}} & \multicolumn{2}{c|}{DTB70} & \multicolumn{2}{c|}{UAVDT} & \multicolumn{2}{c|}{UAV123} & \multicolumn{2}{c|}{UAVTrack112} & \multicolumn{2}{c|}{UAVTrack112$\_$L} & \multicolumn{2}{c|}{Avg.} & \multicolumn{2}{c}{FPS} \\ \cline{3-16}
    && AUC(\%) & P(\%) & AUC(\%) & P(\%) & AUC(\%) & P(\%) & AUC(\%) & P(\%) & AUC(\%) & P(\%) & AUC(\%) & P(\%) & GPU & CPU \\
    \midrule
    SGLATrack-DeiT$^*$                           & Ours     & {\color{red}65.1}  & {\color{red}84.4} & {\color{red}59.9}  & {\color{blue}81.9}  & {\color{red}66.9} & {\color{red}84.9} & {\color{red}67.5} & {\color{red}82.8} & {\color{red}64.0} & 79.2 & {\color{red}64.7} & {\color{red}82.6} & {\color{red}224.7} & 74.8  \\   
    \midrule
    AVTrack-DeiT \cite{AVTrack}     & ICML'24   & {\color{blue}65.0}  & {\color{blue}84.3}    & {\color{blue}58.7}   & {\color{red}82.1}     & {\color{blue}66.8}   & {\color{blue}84.8}       & {\color{blue}65.4}    & 80.3     & {\color{blue}62.7}     & 78.2  & {\color{blue}63.7} & {\color{blue}81.9} & {\color{blue}197.3}  & -   \\
    SMAT \cite{SMAT}           & WACV'24   & 63.8  & 81.9    & 58.7   & 80.8     & 64.6   & 81.8       & 65.3    & {\color{blue}82.2}     & 60.8     & 76.5  & 62.6 & 80.6 & 158.4  & -  \\ 
    HiT-Small \cite{HiT}             & ICCV'23   &58.9  & 74.8   & 46.6   & 61.6     & 63.8   & 80.6       & 63.6    & 78.0     & 59.8     & 74.6  & 58.7 & 74.1 & 194.6  & -   \\
    MVT \cite{MVT}          & BMVC'23      & 60.4  & 79.2    & 47.5   & 65.8     & 59.9   & 78.6       & 60.8    & 78.2     & 46.2     & 59.5  & 55.0 & 72.3 & 175.1  & -    \\
    TCTrack++ \cite{TCTrack++}  & TPAMI'23 & 62.6  & 81.5    & 52.6   & 72.2     & 59.3   & 78.0       & 64.0    & 81.5     & 62.4     & {\color{red}82.0}  & 60.2 & 79.0 & 136.9  & -   \\
    TCTrack \cite{TCTrack}     & CVPR'22   & 62.2  & 81.2    & 53.0   & 72.5     & 60.5   & 80.0       & 59.4    & 76.6     & 58.3     & 78.6  & 58.7 & 77.8 & 135.8  & -    \\
    UDAT \cite{UDAT}           & CVPR'22   & 61.8  & 80.6    & 59.2   & 80.1     & 59.0   & 76.1       & 61.6    & 78.9     & 59.6     & 77.2  & 60.2 & 78.6 & 51.2  & -    \\
    HCAT \cite{HCAT}           & ECCV'22   & 63.7  & 83.5    & 54.3   & 74.2     & 64.8   & 84.2       & 66.0    & 81.9     & 61.7     & 78.2  & 62.1 & 80.4 & 110.1  & -    \\
    LightTrack \cite{LightTrack} & CVPR'21 & 59.1  & 76.4    & 61.2   & 80.4     & 62.7   & 78.3       & 63.8    & 81.1     & 62.0     & {\color{blue}79.9}  & 61.8 & 79.2 & 129.8  & -    \\
    HiFT \cite{HIFT}           & ICCV'21   & 59.4  & 80.2    & 47.5   & 65.2     & 59.0   & 78.7       & 57.0    & 74.2     & 55.1     & 73.4  & 55.6 & 74.3 & 122.6  & - \\
    AutoTrack \cite{Autotrack} & CVPR'20   & 47.8  & 71.6    & 45.0   & 71.8     & 47.2   & 68.9       & 46.5    & 69.4     & 40.2     & 67.5  & 45.3 & 69.8 & -  & 55.8   \\
    ARCF \cite{ARCF}           & ICCV'19   & 47.2  & 69.4    & 45.8   & 72.0     & 46.8   & 67.1       & 45.6    & 67.2     & 39.3     & 64.0  & 44.9 & 67.9 & -  & 57.6   \\   
    STRCF \cite{STRCF}       & CVPR'18   & 43.7  & 64.9    & 41.1  & 62.9     & 48.1  & 68.1          & 43.0    & 63.5     & 36.0     & 60.9  & 42.4 & 64.1 & -  & 27.4   \\
    MCCT-H \cite{MCCT_H}       & CVPR'18   & 40.5  & 60.4    & 40.2   & 66.8     & 45.7   & 65.9       & 43.6    & 63.4     & 37.5     & 58.5  & 41.5 & 63.0 & -  & 57.0   \\
    ECO-HC \cite{ECO}          & CVPR'17   & 44.8  & 63.5    & 41.6   & 69.4     & 49.6   & 71.0       & 47.4    & 68.6     & 41.7     & 64.8  & 45.0 & 67.5 & -  & {\color{blue}79.5}   \\
    fDSST \cite{DSST}         & TPAMI'17  & 35.7  & 53.4    & 38.3   & 66.6     & 40.5   & 58.3        & 39.1    & 56.8     & 30.6     & 49.1  & 36.8 & 56.8 & -  & {\color{red}184.8}  \\
    \midrule 
    \multicolumn{13}{l}{\multirow{1}{*}{\textit{SGLATrack with different backbones}} }  \\
    \midrule
    SGLATrack-EVA & Ours        & 63.8 & 83.5 & 57.9  & 79.9 & 65.1 & 83.05  & 66.9 & 81.3 & \underline{64.7} & \underline{81.7} & 63.7 & 81.9 & \underline{236.9} & \underline{77.2}  \\
    SGLATrack-ViT & Ours        & \underline{65.8} & \underline{84.8} & \underline{60.0}  & \underline{82.4} & 66.1 & 83.9 & 67.3  & 82.6 & 64.3 & 80.5 & \underline{64.7} & \underline{82.8} & 223.2  & 74.2  \\
    SGLATrack-DeiT$^*$ & Ours   & 65.1 & 84.4 & 59.9  & 81.9 & \underline{66.9} & \underline{84.9}  & \underline{67.5} & \underline{82.8}  & 64.0 & 79.2 & \underline{64.7} & 82.6 & 224.7  & 74.8 \\
    \bottomrule
  \end{tabular} }
  \label{tab:all}
\end{table*}

\begin{table*}[t]
\centering
\caption{Comparison with state-of-the-art methods on UAV123@10fps. The top two results are highlighted with {\color{red}red}, {\color{blue}blue}, respectively.}
\resizebox{\textwidth}{!}{
     \fontsize{16}{17}\selectfont
\begin{tabular}{c|cccccccccc|ccc}
\toprule
\multirow{2}{*}{Method} & HiFT & LightTrack & HCAT & UDAT & TCTrack & TCTrack++ & MVT & HiT & SMAT & AVTrack & SGLATrack & SGLATrack & SGLATrack\\

       & \cite{HIFT} & \cite{LightTrack} & \cite{HCAT} & \cite{UDAT} & \cite{TCTrack} & \cite{TCTrack++} & \cite{MVT} & -Small \cite{HiT} & \cite{SMAT} & -DeiT \cite{AVTrack} & -DeiT$^*$ & -ViT & -EVA\\
\midrule
AUC(\%) & 57.0 & 59.9 & 64.4 & 58.5 & 58.7 & 59.9 & 57.6 & 64.3 & 63.5  & {\color{red}65.8} & {\color{blue}65.5} & 64.5 & 64.3 \\

P(\%)   & 74.9 & 75.1 & 81.9 & 77.8 & 77.4 & 78.0 & 73.2 & 80.9 & 80.4 & {\color{red}83.2} & {\color{blue}82.6} & 82.2 & 81.7 \\

FPS & 122.6 & 129.8 & 110.1 & 51.2 & 135.8 & 136.9 & 175 & 194.6 & 158 & 197.3 & {\color{blue}224.7} & 223.2 & {\color{red}{236.9}} \\
\bottomrule
\end{tabular} }
\label{tab:uav10fps}
\end{table*}

\textbf{Training strategy.} Following traditional protocols, we train our models on four datasets, including LaSOT \cite{lasot}, COCO \cite{coco}, TrackingNet \cite{trackingnet}, and GOT-10k \cite{got} (with 1,000 videos removed according to \cite{Stark}). Same as OSTrack \cite{OSTrack}, we use typical commonly used data augmentation methods, such as horizontal flipping and brightness jittering. We train all three trackers uniformly with 300 epochs and 60k matching pairs per epoch. With AdamW \cite{Adamw} optimizer, the learning rates are set to 4 × $10^{-4}$ for the prediction head, and 4 × $10^{-5}$ for the backbone and the selection module. The learning rates drop by a factor of 10 after 240 epochs. The model training is conducted on a server with four 80GB Tesla A100 GPUs, and the batch size is uniformly fixed at 32.

\textbf{Inference.} We follow the common practice \cite{transt,OSTrack,OCEAN} and employ the Hamming window to incorporate positional priors. Specifically, a penalty is applied to the classification map $\textbf{p}$ using the Hanning window. After that, the box corresponding to the highest score is selected as the prediction bounding box.

\subsection{Results and Comparisons}
\textbf{Datasets.} To demonstrate the effectiveness of our method, six datasets are used for evaluation, including  DTB70 \cite{DTB70}, UAVDT \cite{UAVDT}, UAV123 \cite{uav123}, UAV123@10fps \cite{uav123}, UAVTrack112 \cite{UAV112}, and  UAVTrack112$\_$L \cite{UAV112}.
Specifically, DTB70 consists of 70 UAV sequences, encompassing various challenges such as UAV motion, cluttered scenes and objects with different sizes. UAVDT is primarily employed for evaluating vehicle tracking under situations of different weather conditions, flying altitudes, and camera vies. UAV123 is a large-scale UAV tracking benchmark including 123 sequences with more than 112K frames. UAV123@10fps is constructed by down-sampling the original UAV123 benchmark from 30 FPS to 10 FPS to study the impact of camera capture speed on tracking performance. UAVTrack112 comprises 112 challenging aerial scenarios, totaling over 100K frames. UAVTrack112$\_$L serves as the biggest long-term aerial tracking benchmark currently, including over 60k frames.

\textbf{Performance evaluation.}
We compare the SGLATrack variant with the best overall performance, i.e., SGLATrack-DeiT$^*$, against the current state-of-the-art (SOTA) lightweight trackers. \cref{tab:all} shows the AUC and precision of all the compared trackers on  DTB70, UAVDT, UAV123, UAVTrack112 and UAVTrack112$\_$L datasets. As can be seen, our SGLATrack-DeiT$^*$ achieves new AUC scores on the DTB70, UAV123, UAVTrack112, and UAVTrack112$\_$L datasets, and maintains top-three precision across all datasets. Particularly on UAVTrack112, SGLATrack-DeiT$^*$ outperforms the second-best AVTrack by 2.1\% in AUC and 2.5\% in precision. An interesting observation is that SGLATrack-Deit$^*$ achieves the best AUC scores on UAVDT and UAVTrack112$\_$L, yet its precision is inferior to some trackers with much lower AUC scores than ours. We speculate that this discrepancy arises from the bias introduced by the limited number of sequences in these datasets, as UAVDT and UAVTrack112$\_$L contain only 50 and 45 sequences respectively. Besides, the comparison results on UAV123@10fps are shown in \cref{tab:uav10fps}. Although the performance of SGLATrack-DeiT$^*$ is inferior to AVTrack, it significantly outperforms AVTrack in inference speed. The outstanding performance on the six datasets indicates that our approach can accurately disable redundant layers without notable performance drops, thereby confirming our intuition about the layer redundancy of lightweight ViTs. We also compare the overall performance of different SGLATrack variants. It can be seen that SGLATrack-ViT and SGLATrack-DeiT$^*$ achieve competitive performance, which demonstrates the generalization ability of our similarity-guided layer adaptation approach.

\textbf{Speed evaluation.}  For a fair comparison, the experiments of inference speeds are conducted on a same PC machine with an Intel i7-9700KF CPU and an Nvidia GTX 2080Ti GPU. As demonstrated in \cref{tab:all}, our SGLATrack family achieves SOTA GPU speeds, with SGLATrack-ViT, SGLATrack-DeiT$^*$, and SGLATrack-EVA running at effective speeds of 223.2 FPS, 224.7 FPS, and 236.9 FPS, respectively. Moreover, our SGLATrack family also demonstrates competitive real-time CPU speeds, even surpassing some famous DCF-based trackers such as AutoTrack \cite{Autotrack} and ARCF \cite{ARCF}. The comparison results of inference speed indicate that our approach has strong acceleration capabilities, making it highly suitable for ViT-based UAV tracking.

\subsection{Ablation Study and Analysis}
To demonstrate the effectiveness of proposed layer adaptation (LA) approach, we conduct extensive ablation studies on two large-scale datasets, i.e., UAV123 \cite{uav123} and UAVTrack112 \cite{UAV112}. Default settings are marked in \colorbox{gray!25}{gray}.

%%%%%%%%%%%%%%%%%%%%%%%%%%%%%%%%%%%%%%%%%%%%%

% \begin{table}[t]
% \centering
% \caption{Ablation studies of incorporating LA into different backbones. Note that tracking performance is shown in AUC.}
% \label{tab:backbone}
%  \resizebox{\linewidth}{!}{
%  \fontsize{8}{9}\selectfont
% \begin{tabular}{ccccccc}
% \toprule
%         Variant                     & LA & \multicolumn{2}{c}{UAV123}       & \multicolumn{2}{c}{UAVTrack112}  & FPS \\ \midrule
% \multirow{2}{*}{SGLATrack-EVA}  &    & \multicolumn{2}{c}{65.3} & \multicolumn{2}{c}{67.1} & 185.4 \\
%                                 & \checkmark  & \multicolumn{2}{c}{65.1} & \multicolumn{2}{c}{66.9} & 236.9 \\ \midrule
% \multirow{2}{*}{SGLATrack-ViT}  &    & \multicolumn{2}{c}{66.4} & \multicolumn{2}{c}{67.7} & 172.8 \\
%                                 & \checkmark  & \multicolumn{2}{c}{66.1} & \multicolumn{2}{c}{67.3} & 223.2 \\ \midrule
% \multirow{2}{*}{SGLATrack-DeiT$^*$} &    & \multicolumn{2}{c}{67.1} & \multicolumn{2}{c}{67.8} & 175.5 \\
%                                 & \checkmark  & \multicolumn{2}{c}{66.9} & \multicolumn{2}{c}{67.5} & 224.7 \\ \bottomrule
% \end{tabular}}
% \end{table}

\begin{table}[t]
\centering
\caption{Ablation studies of determining saturated layers with different methods. Note that tracking performance is shown in AUC.}
\label{tab:backbone}
 \resizebox{\linewidth}{!}{
 \fontsize{15}{16}\selectfont

\begin{tabular}{ccccccccc}

\toprule
        Variant                     & LA & \multicolumn{2}{c}{UAV123}       & \multicolumn{2}{c}{UAVTrack112}  & FPS & Params. & FLOPs \\ 
        \midrule
\multirow{2}{*}{SGLATrack-EVA}  &    & \multicolumn{2}{c}{65.3} & \multicolumn{2}{c}{67.1} & 185.4 & 5.76 & 1.73 \\
                                & \checkmark  & \multicolumn{2}{c}{65.1} & \multicolumn{2}{c}{66.9} & 236.9 & 4.15 & 1.20 \\ \midrule
\multirow{2}{*}{SGLATrack-ViT}  &    & \multicolumn{2}{c}{66.4} & \multicolumn{2}{c}{67.7} & 172.8 & 7.98 & 2.39 \\
                                & \checkmark  & \multicolumn{2}{c}{66.1} & \multicolumn{2}{c}{67.3} & 223.2 & 5.81 & 1.68 \\ \midrule
\multirow{2}{*}{SGLATrack-DeiT$^*$} &    & \multicolumn{2}{c}{67.1} & \multicolumn{2}{c}{67.8} & 175.5 & 7.98 & 2.39 \\
                                & \checkmark  & \multicolumn{2}{c}{66.9} & \multicolumn{2}{c}{67.5} & 224.7 & 5.81 & 1.68 \\ 
                                \bottomrule
\end{tabular}
}
% \vspace{-2.3em}
\end{table}

%%%%%%%%%

\begin{table}[t]
\centering
\caption{Ablation studies of determining saturated layers with different methods. Note that tracking performance is shown in AUC.}
\label{tab:saturated}
 \resizebox{\linewidth}{!}{
 \fontsize{5}{6}\selectfont
\begin{tabular}{ccclclc}
\hline
Method                    & Value & \multicolumn{2}{c}{UAV123} & \multicolumn{2}{c}{UAVTrack112} & FPS   \\ \hline
\multirow{3}{*}{Adaptive} & $\mu$ = 0.88  & \multicolumn{2}{c}{65.4}   & \multicolumn{2}{c}{65.9}        & 204.6 \\
                          & $\mu$ = 0.90  & \multicolumn{2}{c}{66.5}   & \multicolumn{2}{c}{66.3}        & 192.8 \\
                          & $\mu$ = 0.92  & \multicolumn{2}{c}{67.0}   & \multicolumn{2}{c}{67.7}        & 184.1 \\ \hline
                
                          & $l^*$ = 5     & \multicolumn{2}{c}{66.1}   & \multicolumn{2}{c}{66.4}        & 239.6     \\ 
\rowcolor{gray!25}\textbf{Direct}  &  \textbf{$\bm{l^*}$ = 6 }    & \multicolumn{2}{c}{66.9}   & \multicolumn{2}{c}{67.5}        & 224.7  \\ 
                          & $l^*$ = 7     & \multicolumn{2}{c}{66.9}   & \multicolumn{2}{c}{67.7}        & 211.3    \\
                          \hline
\end{tabular} }
\end{table}

%%%%%%%%%

\begin{table}[t]
\centering
\caption{Ablation studies of the optimization of selection module. Note that tracking performance is shown in AUC.}
\label{tab:selection_module}
 \resizebox{\linewidth}{!}{
 \fontsize{5}{6}\selectfont
\begin{tabular}{lcclclc}
\hline
\# & Method           & \multicolumn{2}{c}{UAV123} & \multicolumn{2}{c}{UAVTrack112} & FPS   \\ \hline
1  & Fixed layer      & \multicolumn{2}{c}{65.4}   & \multicolumn{2}{c}{65.8}        & 232.9 \\  \rowcolor{gray!25}
\textbf{2}  &\textbf{Maximizing}        & \multicolumn{2}{c}{66.9}   & \multicolumn{2}{c}{67.7}        & 224.7 \\
3  & Minimizing       & \multicolumn{2}{c}{65.7}   & \multicolumn{2}{c}{64.5}        & -     \\
4  & Random selection & \multicolumn{2}{c}{66.0}   & \multicolumn{2}{c}{66.2}        & -     \\ \hline
\end{tabular} }
\end{table}

\textbf{LA on different backbones.} Our LA approach can be seamlessly integrated into existing ViTs and adaptively optimize their redundant layers. As shown in \cref{tab:backbone}, we incorporate LA into three different ViTs to study its impact on tracking performance. It can be observed that our LA approach significantly improves inference speed with only a slight performance drop. In the best case, the approach only results in a 0.2\% AUC decrease, such as SGLATrack-EVA on UAV123 and UAVTrack112. In the worst case, the AUC loss is just 0.4\% (SGLATrack-ViT on UAVTrack112), which is considered acceptable given the nearly 22\% increase in speed, 27\% reduction in parameters, and 30\% decrease in Flops. These results provide strong evidence for the generalizability of proposed LA approach.

\textbf{Study on saturated layer.} The saturated layer $l^*$ is crucial for balancing precision and speed in our framework. We investigate two different methods to determine when saturation state is reached. The first method calculates the cosine similarity between search features layer by layer during inference to adaptively determine the saturated layer. Given a certain threshold $\mu$, if the cosine similarity exceeds this threshold, it indicates that saturation has been reached. The second method involves using prior knowledge to directly set a fixed value for the saturated layer $l^*$, making it simpler and more effective. \cref{tab:saturated} shows the impact of SGLATrack-DeiT$^*$ using the two method on performance and speed. As can be seen, the adaptive method minimizes performance drop when the threshold is set to 0.92. The direct method achieves the best accuracy-speed trade-off when $l^*$ is set to 6 with a precision that is very close to that of the adaptive method, but with a significantly higher speed. We suppose that there are three main reasons for this result: 1) The direct method does not require calculating the cosine similarity layer-by-layer, thus avoiding additional time costs. 2) Directly setting a fixed value for $l^*$ forces the model to accelerate its ponder and reach saturation stats earlier. In other words, if $l^*$ is set to 6, the model might already saturate before the 6-th layer. 3) Our proposed selection module can effectively mitigate the precision drop when a large number of layers are deactivated. Hence, the direct method with $l^*$ being set to 6 is adopted.

\textbf{Study on selection module.} 
The selection module compensates for precision drop by dynamically retaining an optimal ($l^{*}+k$)-th layer. Here, we provide various experiments to study the impact of this module. \#1 Fixed layer. $k$ is set to 1, meaning the first layer after the saturation layer $l^*$ is retained. Since $k$ is fixed, the selection module is removed in this experiment. \#2 Maximizing. Encourage the model to further strengthen its focus on the target by maximizing the cosine similarity based on \cref{eq:objective}. \#3 Minimizing. By minimizing cosine similarity, the model is encouraged to distract its attention to continue mining scene relationships. \#4 Random selection. Unlike \#2 and \#3, no supervision is applied to this module. Note that experiment \#2, \#3, and \#4 all adopt the selection module, so they share the same speed. \cref{tab:selection_module} shows the impact of different methods on performance and speed. The maximizing-based method \#2 achieves the best precision-speed trade-off, verifying the effectiveness of our layer-wise similarity loss. Moreover, comparing experiments \#1 and \#2, the method in \#2 shows a slight speed decrease, due to the introduced computational cost of the selection module.

% comparing experiments \#1 with \#2, the method in experiment \#2 shows only a slight decrease in speed, which is due to the computational cost introduced by our selection module. 

\begin{figure}[t]
  \centering
   \includegraphics[width=3.2in]{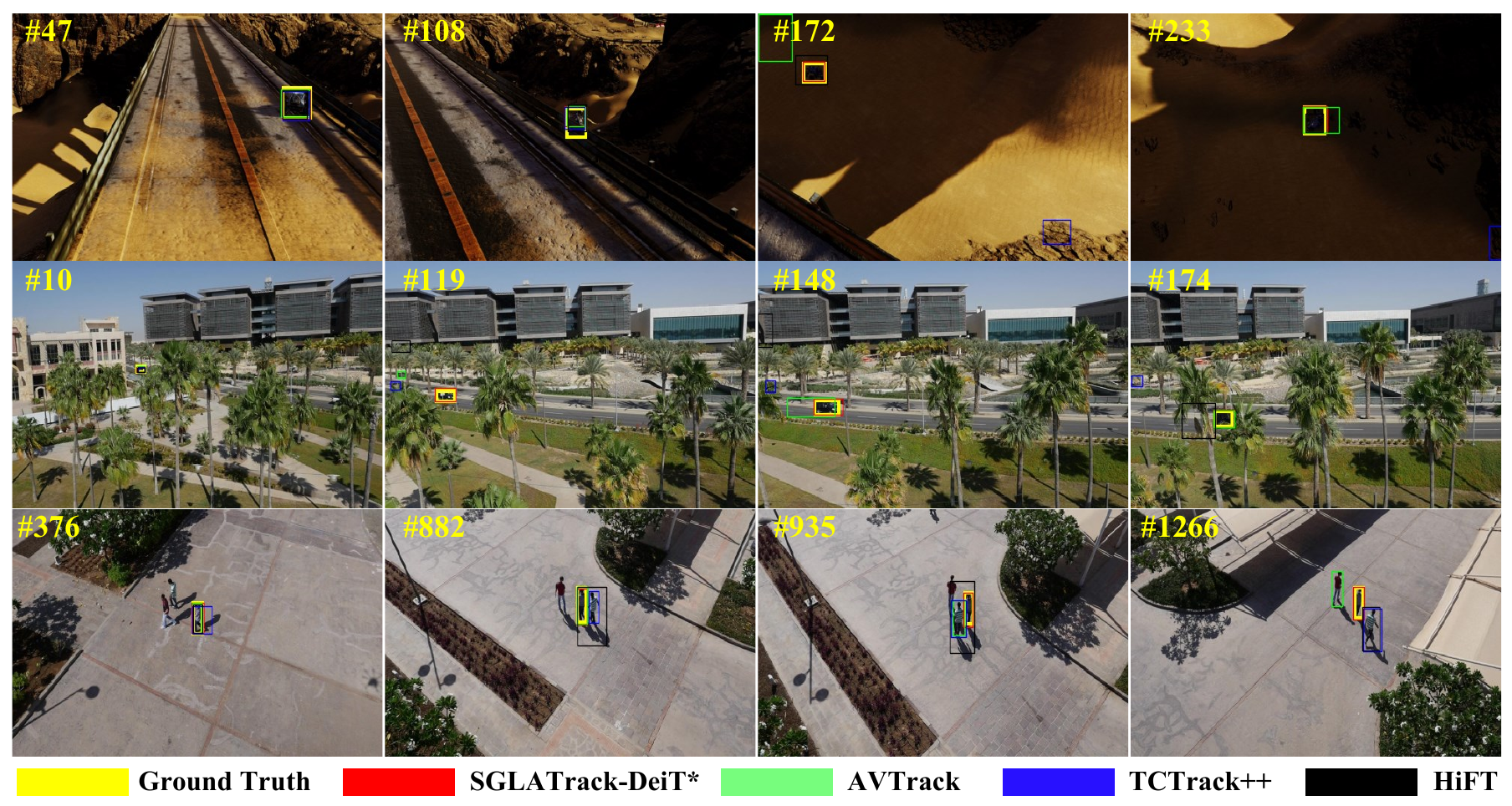}
   \caption{Qualitative comparisons of our tracker against other three SOTA trackers. Best viewed in color and by zooming in.}
   \label{fig:quli}
\end{figure}

\begin{figure}[t]
  \centering
   \includegraphics[width=2.7in]{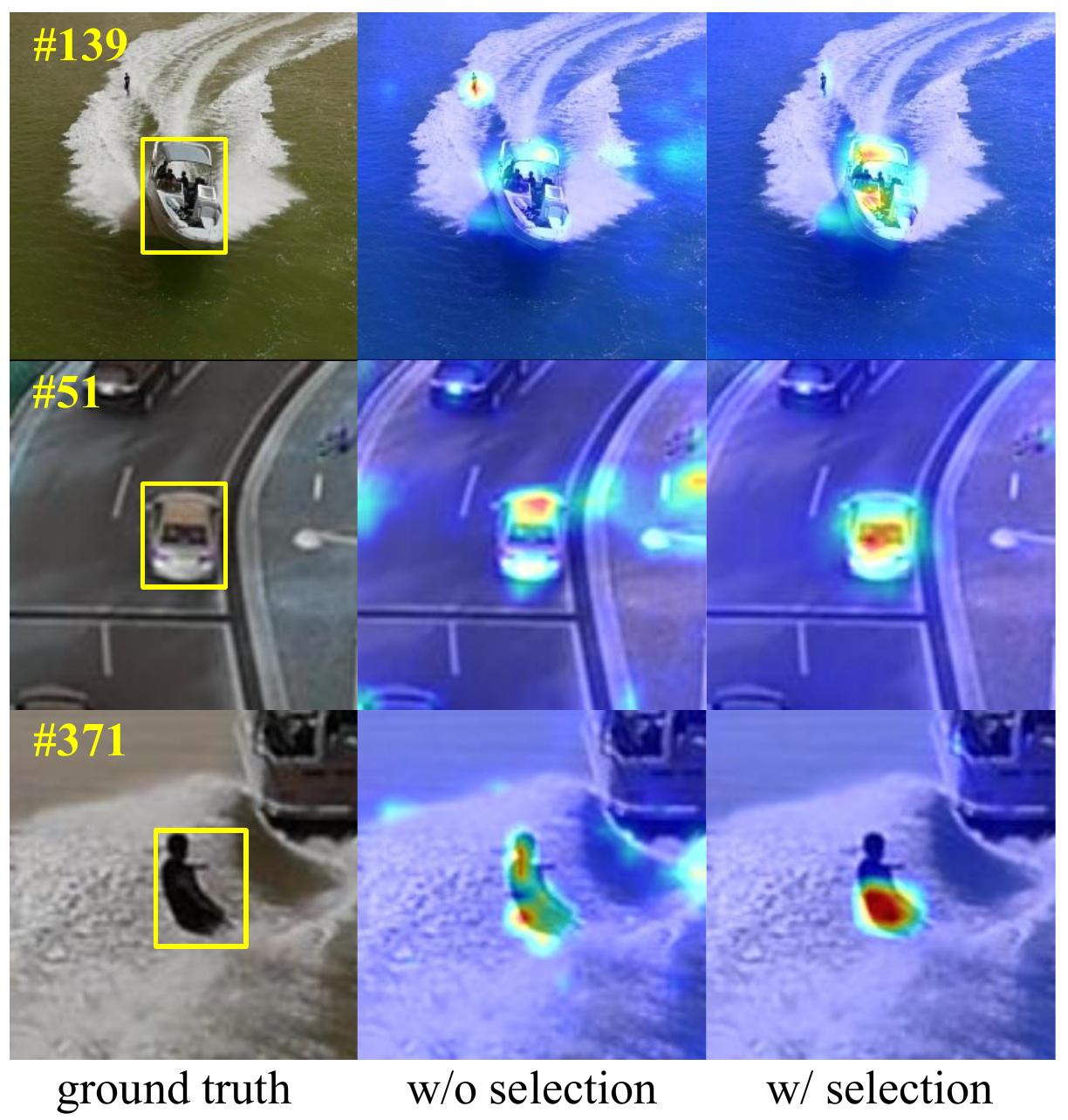}
   \caption{Comparison of attention maps. Note that w/o and w/ denote the tracker without and with selection module, respectively.}
   \label{fig:attn}
\end{figure}

\subsection{Qualitative Comparison and Visualization}
\textbf{Qualitative results.} To intuitively demonstrate the effectiveness of proposed approach, we provide some qualitative comparison results in UAV123 \cite{uav123}. The compared trackers include SGLATrack-DeiT$^*$, AVTrack \cite{AVTrack}, TCTrack++ \cite{TCTrack++}, HiFT \cite{HIFT}. As shown in \cref{fig:quli}, in some challenging scenarios, our SGLATrack-DeiT$^*$ with a large number of disabled layers can still maintain robust tracking performance and even outperform other trackers. These qualitative results provide additional evidence of the effectiveness of our similarity-guided layer adaptation approach.

\textbf{Visualization of attention.} To intuitively demonstrate the effectiveness of layer-wise similarity loss, we visualize the cross-attention maps between the center part of template region and the search region. More specifically, we provide the cross-attention maps of the model with selection module and the model without selection module. Both models use the same number of VIT layers, but the last layer of the model without the selection module is $l^*+1$, while the last layer of the model with the selection module is $l^*+k$. As shown in \cref{fig:attn}, the attention in the $(l^*+k)$-th layer is more focused on the target compared to the $(l^*+1)$-th layer, suggesting that dynamically retaining the layer under the condition of equal layers contributes to target recognition. These visualization examples verify that our layer-wise similarity loss effectively optimizes the selection module, allowing the model to maximize its attention on the target in situations where a large number of layers are disabled.

\subsection{Real-world Test}
To validate the tracking efficiency under real-world conditions, we deploy SGLATrack-DeiT$^*$ on an embedded platform, NVIDIA Jetson TX2 4GB. Note that the computational performance of the Jetson TX2 is far inferior to that of the Jetson AGX Xavier and Jetson AGX Orin, yet our tracker is still able to run at a real-time speed of 33 FPS without using any acceleration techniques such as TensorRT. Therefore, the real-world tests on the embedded system directly validate the practicability of our trackers in real-world applications.
\section{Conclusion}
In this paper, we make the first attempt to explore the layer redundancy in tiny ViTs, from the perspective of layer-by-layer feature changes and result variations. Based on the analysis, we propose a selection module that disables many redundancy layers and selectively retains an optimal subsequent layer to balance the accuracy-speed trade-off. This module is optimized via a layer-wise similarity loss, which allows the model to maximize its focus on targets in situations where a large number of layers are disabled. Building upon this module, we tailor previously one-stream architecture into an efficient similarity-guided layer-adaptive tracking framework for real-time UAV tracking. Experimental results show that our tracker achieves state-of-the-art performance and tracking speed on six benchmarks. 

% We hope that our findings can contribute to more advanced design guidelines for accelerating one-stream ViT-based UAV tracking.

% We hope that our findings can provide valuable insights and contribute to more advanced design guidelines for accelerating one-stream ViT-based UAV tracking.

% 

\section{Acknowledgement}
This work is supported by the Project of Guangxi Science and Technology (No.2024GXNSFGA010001), the National Natural Science Foundation of China (No.U23A20383, 62472109 and 62466051), the Guangxi "Young Bagui Scholar" Teams for Innovation and Research Project, the Research Project of Guangxi Normal University (No.2024DF001), the Innovation Project of Guangxi Graduate Education (YCBZ2024083), and the grant from Guangxi Colleges and Universities Key Laboratory of Intelligent Software (No.2024B01).

% This paper primarily explores the layer redundancy in tiny ViTs and proposes a selection module to tailor a highly efficient family of trackers. Although this module successfully accelerates Tiny ViT-based UAV tracking, the layer redundancy in other versions, such as Small or Base ViTs, has not been explored. In the future, we believe that with advancements in hardware, the Small or Base versions of ViT can also be deployed on UAVs. To this end, a promising direction is to explore the redundancy in the Small and even Base versions of ViT, and to design corresponding lightweight solutions for them to achieve a better balance between accuracy and speed.

% Moreover, in the future, we consider extending our module to image classification tasks whose backbones are similar to that of tracking tasks.

% aiming to provide more advanced guidelines for 

{
    \small
    \bibliographystyle{ieeenat_fullname}
    \bibliography{main}
}

% WARNING: do not forget to delete the supplementary pages from your submission 
% \input{sec/X_suppl}

\end{document}